\documentclass[12pt]{article}

\usepackage[margin=1.2in]{geometry}
\usepackage{amsmath,amsthm,amssymb}
\usepackage{hyperref}
\usepackage{physics}
\usepackage{enumitem}
\usepackage{graphicx}
\usepackage{subcaption}
\usepackage{cleveref}
\usepackage[svgnames,dvipsnames,x11names,table]{xcolor}
\usepackage{tikz}
\usepackage{pgfplots}
\usetikzlibrary{decorations.markings}
\usetikzlibrary{shapes.misc}
\usetikzlibrary{calc}
\usetikzlibrary{arrows}
\usepackage{subcaption}
\usepackage{cancel}
\usepackage{mathtools}
\usepackage{epigraph}
\usepackage{bm}
\usepackage{tensor}
\usepackage{setspace}
\usepackage{tcolorbox}
\usepackage[numbers]{natbib}
\usepackage[linesnumbered,ruled,vlined]{algorithm2e}
\usepackage{multicol}  
\usepackage{dsfont}

\usepackage[framemethod=default]{mdframed}
\newmdenv[skipabove=7pt,
skipbelow=7pt,
rightline=false,
leftline=false,
topline=false,
bottomline=false,
backgroundcolor=gray!10, 
linecolor=gray,
innerleftmargin=5pt,
innerrightmargin=5pt,
innertopmargin=5pt,
innerbottommargin=5pt,
leftmargin=0cm,
rightmargin=0cm,
linewidth=4pt]{eBox}

\hypersetup{
    colorlinks=true,
    linkcolor={red!50!black},
    citecolor={blue!50!black},
    urlcolor={blue!80!black}
}

\numberwithin{equation}{section}

\crefname{table}{Table}{Tables}
\crefname{equation}{Eq.}{Eqs.}
\crefname{appendix}{App.}{Apps.}
\crefname{section}{Sec.}{Secs.}
\crefname{figure}{Fig.}{Figs.}

\def \bplus{{b_{+}}}
\def \bminus{{b_{-}}}
\def \d{{\rm d}}
\def \e{{\bm e}} 
\def \g{{\sigma}}

\def \D{{\mathrm D}}

\def \K{{\arr{\P}}}
\def \O{{\mathcal O}}
\def \P{{p}}

\def \s{s}
\def \t{t}
\def \th{{\bm{\theta}}}

\def \B{{B}} 
\def \E{{\mathbb{E}}}

\def \T{{T}}

\def \N{{\mathcal N}}

\def \pd{{\partial}}
\def \ep{{\epsilon}}
\def \Dim{{D}}

\def \deq {{\, := \,}}

\def \nn {\nonumber}

\newcommand{\arr}[1]{{\stackrel{\leftarrow}{#1}}}

\def \inP {{\P_{0}}}
\def \finP {{\P_{\rm d}}}
\def \finy {{y_{\rm d}}}

\newcommand{\AP}[1]{{}}

\begin{document}

\begin{titlepage}
\setcounter{page}{1} \baselineskip=15.5pt
\thispagestyle{empty}
\vskip 60 pt

\begin{center}
{\fontsize{18}{18} \bf Diffusion Density Estimators}
\end{center}

\vskip 20pt
\begin{center}
{\fontsize{12}{18}\selectfont  Akhil Premkumar}
\end{center}

\begin{center}
\vskip 4pt
\textit{{\small Kavli Institute for Cosmological Physics, University of Chicago, IL 60637, USA}
}

\end{center}

\vspace{0.4cm}
 \begin{center}{\bf Abstract}
 \end{center}

\noindent
We investigate the use of diffusion models as neural density estimators. The current approach to this problem involves converting the generative process to a smooth flow, known as the Probability Flow ODE. The log density at a given sample can be obtained by solving the ODE with a black-box solver. We introduce a new, highly parallelizable method that computes log densities without the need to solve a flow. Our approach is based on estimating a path integral by Monte Carlo, in a manner identical to the simulation-free training of diffusion models. We also study how different training parameters affect the accuracy of the density calculation, and offer insights into how these models can be made more scalable and efficient.

\end{titlepage}

\setcounter{page}{2}

\restoregeometry

\begin{spacing}{1.2}
\newpage
\setcounter{tocdepth}{2}
\tableofcontents
\end{spacing}

\setstretch{1.1}
\newpage

\section{Introduction}

Given $N$ data vectors $\{ \finy\}$, what is the underlying distribution $\finP$ from which they were sampled? There is no unique solution for finite $N$, since there are several candidate distributions that could have produced the same $\{ \finy \}$ \cite{Bialek96}. However, intuitively it is clear that the space of admissible solutions must shrink if $N$ is large. Under this assumption, we can construct a reasonable approximation to $\finP$ using \textit{neural density estimators}.

A widely used approach to the problem leverages a neural network to learn an invertible transformation that maps a simple base distribution to a complex target density, a method known as a normalizing flow. Several versions of this approach have found great success as a tool for exact density evaluation \cite{Dinh17,Papamakarios17,Grathwohl19}. A unifying feature of all flow-based techniques is the need to compute a Jacobian determinant to evaluate the density. This calculation can be expensive, especially at higher dimensions, so much effort has been invested in designing architectures and strategies to ameliorate this cost.

Diffusion models have gained significant popularity for modeling high-dimensional data distributions with remarkable fidelity. They are often easier to train than normalizing flows, and produce much higher quality samples. However, diffusion models are predominantly used as generative tools rather than density estimators. In this paper, we explore the latter possibility, study the strengths and weaknesses of diffusion models as density estimators, and probe their sensitivity to various training parameters.
In doing so, we gain valuable insight into how these models can be improved more generally, both in terms of compute/energy efficiency and scalability.

The standard procedure for obtaining density from diffusion models is to convert the generative process into a smooth flow \cite{Song21}. The resulting expressions constitute a neural ODE \cite{Chen18,Grathwohl19}, which can be solved to determine the log density. We review this technique in \cref{sec:ProbabilityFlowODE}. In \cref{sec:DensityEstimation} we introduce a novel approach for density estimation, based on evaluating a path integral. This method closely mirrors the training process of diffusion models, most notably by the fact that it obviates the need to solve the time evolution of the sample of interest. As such, our approach translates the key strengths of diffusion model training, like ease of parallelization and fast convergence, to the problem of evaluating densities.
\begin{figure}
    \centering
    \includegraphics[width=\linewidth]{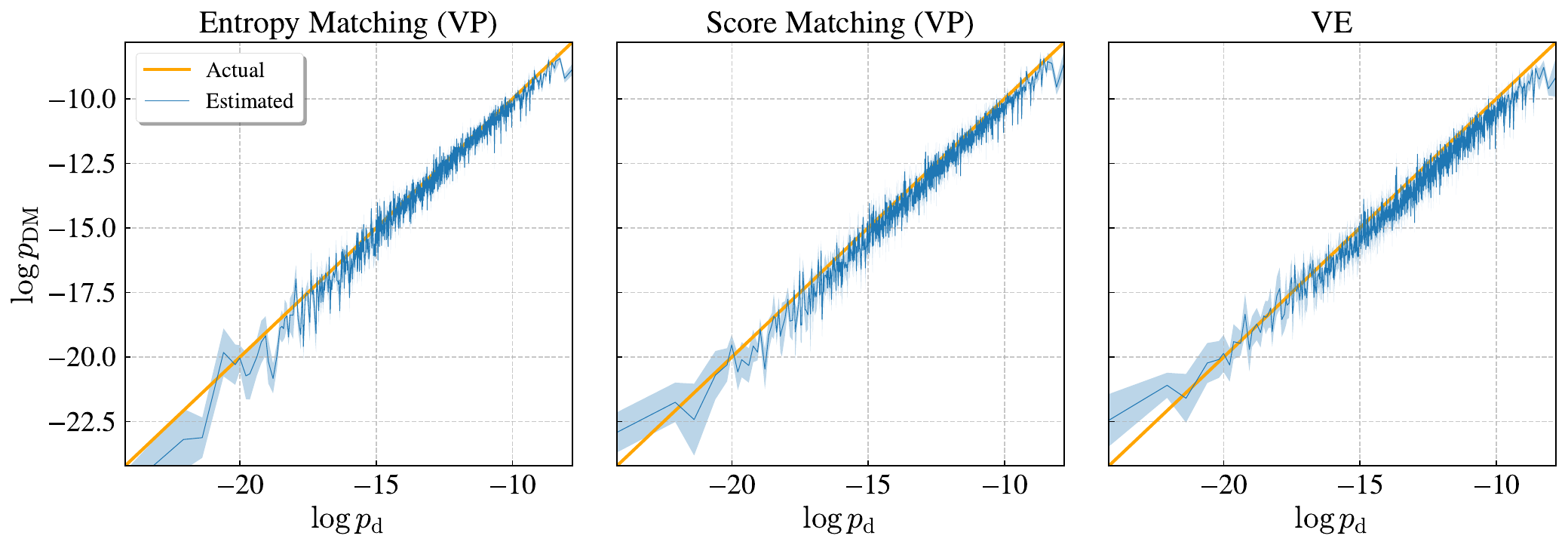}
    \caption{\label{fig:Fast_Likelihoods} Log densities computed from diffusion models using the path integral approach (see \cref{sec:DensityEstimation}). $\finP$ is the true data distribution, a mixture of six Gaussians in $\Dim=9$ dimensions, and $\P_{\rm DM}$ is the density computed from the diffusion model. All models were trained with parameters $N = 8192, n_{\rm t} = 10, n_{\rm ep} = 200$ (see \cref{sec:Experiments}).}
\end{figure}





\paragraph{Notation:} We use the time variable $\s$ for the forward diffusion process, which runs from right ($\s=0$) to left ($\s=T$) in \cref{fig:ReverseDiffusion}. Sometimes we indicate functions of $\s$ as $\arr{f}$ to remove ambiguity when the same function is also expressed in terms of time variable $\t = \T - \s$. That is, $\arr{f}(\s) = \arr{f}(\T-\t) = f(t)$. $\hat{\B}_\s$ and $\B_\t$ denote the Brownian motions associated with the forward and reverse/controlled SDEs, respectively. $\nabla$ is the gradient with respect the spatial coordinates, and $\pd_t, \pd_s$ are partial time derivatives. $\log$ is the natural logarithm. $\finP$ and $\inP$ denote the initial ($\s=0$) and final ($\s=T$) densities for the forward process. $\P_u(\cdot, 0)$ and $\P_u(\cdot, \T)$ are the initial ($\t=0$) and final ($\t=\T$) densities of the controlled process. There is a slight abuse of notation here because $\P_u(\cdot, 0)$ is a prior that does not depend on the control $u$. $\P_u(\cdot, \T)$ is the same as $\P_{\rm DM}(\cdot)$ in \cref{fig:Fast_Likelihoods}.

\section{Diffusion Models}

\subsection{Reverse Diffusion}
\label{sec:ReverseDiffusion}

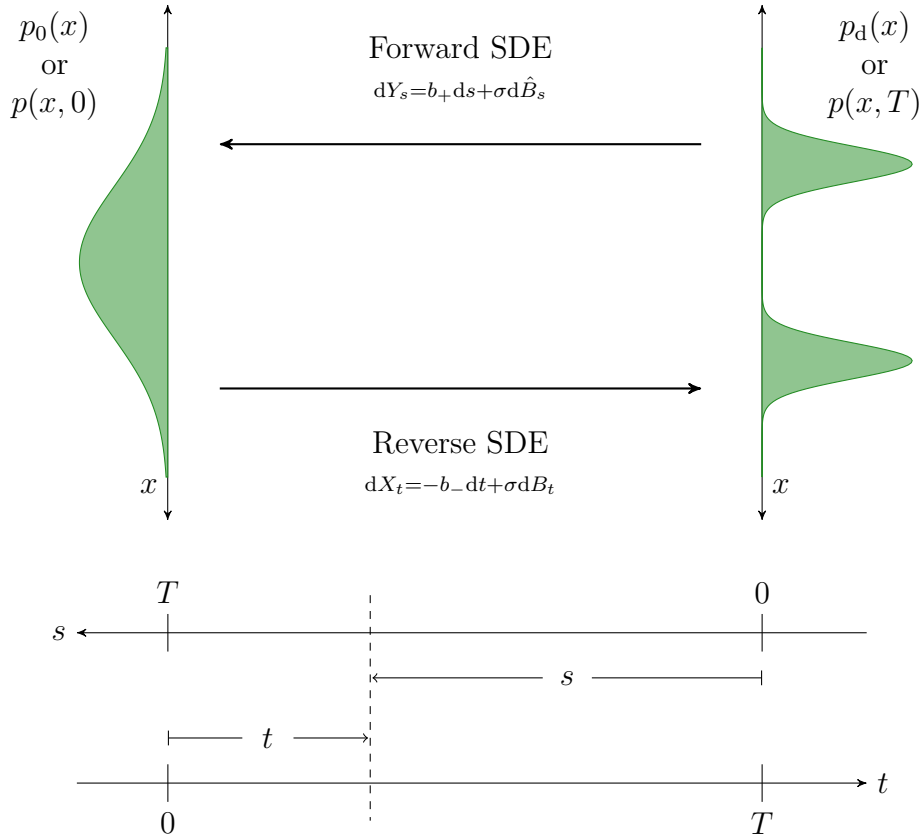
\begin{figure}[h]
    \pgfmathdeclarefunction{gauss}{2}{%
      \pgfmathparse{1/(#2*sqrt(2*pi))*exp(-((x-#1)^2)/(2*#2^2))}%
    }
    
    \centering
    \begin{tikzpicture}
    \begin{scope}[shift={(-5,0)}]
        \begin{axis}[grid=none,
            ymax=2.1,
              axis lines=middle,
              y=1cm,
            y axis line style={draw=none},
            x axis line style={{stealth'}-{stealth'}},
            ytick=\empty,
            xtick=\empty,
            enlargelimits,
            rotate=90,
            ]
            \addplot[ForestGreen,fill=ForestGreen!50,domain=-2.4:2.4,samples=200]  {2.5*gauss(0,0.85)} \closedcycle;
    
            \node[align=center] at (axis cs:2.2,1.5) {$\P_0(x)$ \\ or \\ $\P(x, 0)$};
            \node at (axis cs:-2.5,0.25) {$x$};
        \end{axis}
    \end{scope}
    
    \draw[thick,-{stealth'}] (4.4,5) -- ++(-6.4,0) coordinate[pos=0.5] (a);
    \node[align=center,shift={(0,1)}] at (a) {Forward SDE \\ $ \scriptstyle \d Y_{\s} = \bplus \d \s + \g \d \hat{\B}_{\s}$};
    
    \draw[thick,-{stealth'}] (-2,1.75) -- ++(6.4,0) coordinate[pos=0.5] (b);
    \node[align=center,shift={(0,-1)}] at (b) {Reverse SDE \\ $ \scriptstyle \d X_{\t} = -\bminus \d \t + \g \d \B_{\t}$};

    
    \begin{scope}[shift={(5,0)}]
        \begin{axis}[grid=none,
            ymax=2.1,
            axis lines=middle,
              y=1cm,
            y axis line style={draw=none},
            x axis line style={{stealth'}-{stealth'}},
            ytick=\empty,
            xtick=\empty,
            enlargelimits,
            rotate=-90,
            ]
            \addplot[ForestGreen,fill=ForestGreen!50,domain=-2.4:2.4,samples=200] {gauss(-1.1,0.2) + gauss(1.1,0.2)} \closedcycle;
    
            \node[align=center] at (axis cs:-2.2,1.5) {$\P_{\d}(x)$ \\ or \\ $\P(x,T)$};
            \node at (axis cs:2.5,0.25) {$x$};
        \end{axis}
    \end{scope}

    \begin{scope}[shift={(0,-1.5)}]
        \draw[-{stealth'}] (6.6,0) -- (-3.9,0) node[left] {$\s$};
        \draw (-2.69,0.25) node[above] {$T$} -- ++(0,-0.5);
        \draw (5.21,0.25) node[above] {$0$} -- ++(0,-0.5);
        \draw[|->] (5.21,-0.6) -- (0.03,-0.6) node[circle,fill=white,draw=none,pos=0.5] {$\s$};
    \end{scope}

    \draw[dashed] (0,-1) -- (0,-4);

    \begin{scope}[shift={(0,-3.5)}]
        \draw[-{stealth'}] (-3.9,0) -- (6.6,0) node[right] {$\t$};
        \draw (-2.69,0.25) -- ++(0,-0.5) node[below] {$0$};
        \draw (5.21,0.25) -- ++(0,-0.5) node[below] {$T$};
        \draw[|->] (-2.69,0.6) -- (-0.03,0.6) node[circle,fill=white,draw=none,pos=0.5] {$\t$};
    \end{scope}
    
    
    \end{tikzpicture}
    \caption{\label{fig:ReverseDiffusion}A schematic of the forward and reverse diffusion processes.}
\end{figure}

Generative modeling seeks to reconstruct a data distribution $\P_{\d}$ from a set of its samples $\{ \finy \}$. Diffusion models convert a tractable initial (prior) distribution $\P_0$ to the final distribution $\P_{\d}$, using information gathered from the evolution of $\{ \finy \}$ to a nearly $\P_0$ shaped distribution under a diffusive process. The latter is described by the stochastic differential equation (SDE),
\begin{equation}
    \d Y_{\s} = \bplus(Y, \s) \d \s + \g(\s) \d \hat{\B}_{\s} , \label{eq:ForwardSDE}
\end{equation}
where the time variable $s$ runs from $0$ to $T$ (see \cref{fig:ReverseDiffusion}). Let $\K(y, \s)$ be the distribution that interpolates between $\finP$ and $\inP$ under \cref{eq:ForwardSDE}. If we know the \textit{score function} $\nabla \log \K$ this process can be inverted using the reverse diffusion equation \cite{Nelson66,Anderson82,Haussmann86,Follmer88},
\begin{equation}
    \d X_{\t} = - \bminus (X,\T-\t)  \d \t + \g(\T-\t) \d \B_{\t} , \label{eq:ReverseSDE}
\end{equation}
where $\t = \T - \s$ and the drift term is
\begin{equation}
    \bminus = \bplus(X, \s) - \g^2(\s) \nabla \log \K(X, \s) .
\end{equation}
%
%

Under \cref{eq:ReverseSDE}, the distribution $\P(x,\t)$ that bridges the evolution of $\inP$ back to $\finP$ will be a playback of the forward process in \cref{eq:ForwardSDE}. That is, $\P(x,\t) = \K(x,\T-\t)$. Suppose we replace $\bminus$ in \cref{eq:ReverseSDE} with a different drift term $u$, which we call the \textit{control}. If we start with a distribution $\P_{u}(x, 0)$ close to $\inP(x)$ and evolve by the stochastic process
\begin{equation}
    \d X = - u(X, \t) \d \t + \g(\T-\t) \d \B_\t ,  \label{eq:ControlledSDE}
\end{equation}
the density $\P_u(x,\t)$ of $X_\t$ will differ from $\P(x,\t)$, and land on a terminal distribution $\P_u(x,T) \neq \P(x,\T)$. The Kullback-Leibler divergence between these distributions is bounded as \cite{Premkumar24a}
\begin{equation}
    \int_{0}^{T} \d \t \, \frac{1}{2 \g^2} \E_{\P} \left[ \left\| \bminus - u \right\|^2 \right]
    + \D_{KL} \left ( \inP(\cdot) \| \P_{u}(\cdot, 0) \right)
    \geq
    \D_{KL} \left (\P(\cdot,T) \| \P_u(\cdot, \T) \right) .
    \label{eq:BoundOnKLGeneral}
\end{equation}
In a diffusion model, all or part of $u$ is parameterized by a neural network that is trained to minimize a loss function derived from the first term in the l.h.s. (see \cref{sec:Parameterization}). In this work we consider two forms of $u$,
\begin{equation}
    u =
    \begin{cases}
        \bplus - \g^2 {\bm s}_\th, \quad &\text{Score Matching (SM)} \\
        -\bplus - \g^2 \e_\th, \quad &\text{Entropy Matching (EM)} .
    \end{cases}
\end{equation}
We use the Variance Preserving (VP) process \cite{Sohl-Dickstein15,Song21} for forward diffusion, for which $\bplus(y,\s) = -\beta(\s) y/2$ and $\g(\s) = \sqrt{\beta(\s)}$. In the absence of the drift term, $\bplus$, entropy matching and score matching are identical. The Variance Exploding (VE) process from \cite{Song21}, is an example of a driftless diffusion process. The expectation value in \cref{eq:BoundOnKLGeneral} is taken over stochastic trajectories that take $\finP \to \inP$. In physics, such averages are known as \textit{path integrals}. We discuss them in detail in \cref{sec:PathIntegrals}

\subsection{Probability Flow ODE}
\label{sec:ProbabilityFlowODE}

The standard approach to density estimation with diffusion models entails transforming the model into a continuous flow \cite{Song21}. To understand this, observe that under \cref{eq:ReverseSDE}, the probability distribution evolves according to the Fokker-Planck equation
%
\begin{equation}
    \pd_{\t} \P = \nabla \cdot \left[ (\bplus - \g^2 \nabla \log \P ) \P \right] + \frac{\g^2}{2} \nabla^2 \P .
\end{equation}
This equation can be written in a Liouville form \cite{Maoutsa20}
\begin{equation}
    \pd_{\t} \P = \nabla \cdot f(x,\t) , \qquad f(x,\t) \deq \bplus - \frac{\g^2}{2} \nabla \log \P . 
    \label{eq:ReverseLiouville}
\end{equation}
for the deterministic system described by the ordinary differential equation (ODE)
\begin{equation}
    \frac{\d X}{\d \t} = - f(X,\t) . 
    \label{eq:ReverseODE}
\end{equation}
Furthermore, $\d/\d \t = \pd_{\t} + (\d X/\d \t) \cdot \nabla$, allowing for \cref{eq:ReverseLiouville} to be written as
\begin{equation}
    \frac{\d}{\d t} \log \P = \nabla \cdot f(X,\t) .
    \label{eq:ProbabilisticODE}
\end{equation}
For example, in a score matching model ${\bm s}_\th \approx \nabla \log \P$, so we can replace $\nabla \log \P$ in with ${\bm s}_\th$ to obtain a continuous normalizing flow that approximates \cref{eq:ReverseODE,eq:ProbabilisticODE} \cite{Chen18,Song21}. A point $x_0 \sim \P_0$ flows to $x_\T \sim \P_{\d}$ under \cref{eq:ReverseODE}, and the log likelihood $\log \P_{\d}(x_\T)$ can be determined by solving
\begin{equation}
    \begin{bmatrix}
        x_\T \\
        \log \P_{\d}(x_\T)
    \end{bmatrix} =
    \int_{0}^{\T} \d \t
    \begin{bmatrix}
        - f (x(\t), \t) \\
        \nabla \cdot f(x(\t), \t)
    \end{bmatrix}
    \label{eq:ODESystem}
\end{equation}
with the initial value $\left[x_0,  \log \P_0(x_0) \right]^{\top}$. Note that, to compute the log density at some test point $x_\T$, we must first evolve it to $\t=0$ and then solve the ODE to return it to its original value. The details of solving \cref{eq:ODESystem} are given in \cref{sec:TraceEstimator}. The following practical features of this approach are particularly relevant to this work:
\begin{enumerate}
    \item \label{pt:Sequential} The log likelihood is computed by solving the first order system \cref{eq:ODESystem} using a black box ODE solver. That means $x$ and $\log \P$ at any instant $t$ are determined from their values at the previous instant. Therefore, the calculation is sequential, limiting the speed at which the likelihood can be computed.
    \item Depending on the method of solution, the solver might use adaptive step sizes to control accuracy in regions where the derivative changes rapidly, or when it becomes too small. As a result, the time taken to solve the ODE varies widely between samples (see \cref{fig:Fast_ODEvsPathIntegral}).
    \item \label{pt:Autodiff} The $\nabla \cdot f$ term involves derivatives of the neural network, $\nabla \cdot {\bm s}_\th$ or $\nabla \cdot \e_\th$, which can be expensive to compute especially for higher dimensional $x$. In practice, it is estimated using the Hutchinson trace estimator, along with reverse-mode automatic differentiation \cite{Grathwohl19,Song21}. The estimator introduces fluctuations in the derivative, which in turn affect the final answer.
    %
\end{enumerate}

\section{Path Integrals}
\label{sec:PathIntegrals}

\subsection{Path Averages}
\label{sec:PathAverages}

Under \cref{eq:ForwardSDE}, the probability that a sample $\finy \sim \finP$ travels along a trajectory $y(\s)$ is \cite{Yasue78,Seifert12}
\begin{equation}
    \P[y(\s) | \finy] = \N \exp(- \int_{0}^{\T} \d \s \, \frac{1}{2 \g^2} (\dot{y} - \bplus)^2 + \frac{1}{2} \nabla \cdot \bplus ) . \label{eq:PathProbability}
\end{equation}
We use square brackets to denote functionals over the paths. The path-independent normalization $\N$ ensures that the path weights add up to 1. The expectation value over a path-dependent observable $\O[y(\s)]$ can be computed by integrating it over the weights,
\begin{equation}
    \E_{\K}[\O[y(\s)]] = \int \d \finy \int [\d y(\s)] \, \O[y(\s)] \, \P[y(\s) | \finy] \, \finP(\finy) . \label{eq:PathIntegralAvg}
\end{equation}
Often in our calculations, the observable is an integral over the path,\footnote{Notation: $y(\s)$ denotes the entire path, whereas $y_\s$ is the position at time $\s$.} $\O[y(\s)] = \int_{0}^{\T} \d \s f(y_\s,\s)$, in which case we can compute \cref{eq:PathIntegralAvg} with the regular integral
\begin{equation}
    \E_{\K}[\O[y(\s)]]
    = \int_{0}^{\T} \d \s \, \E[f(y_\s, \s)]
    =  \int_{0}^{\T} \d \s \int \d \finy \int \d y_\s f(y_\s,\s) \P(y_\s, \s | \finy, 0) \finP(\finy) ,
    \label{eq:PathIntegralTransition}
\end{equation}
where we $\P(y_\s,\s | \finy, 0)$ is the finite time transition probability that aggregates the weight of all paths connecting $\finy$ at time $0$ to $y_\s$ at time $\s$. By changing the order of time integration and spatial averaging we are slicing the path integral in a different way: rather than evaluating the time integral over each stochastic trajectory and taking the average, we aggregate the expectation values of integrand $f$ at each time slice, for each $y_\d$.
\begin{figure}
    \centering
    \includegraphics[width=0.8\linewidth]{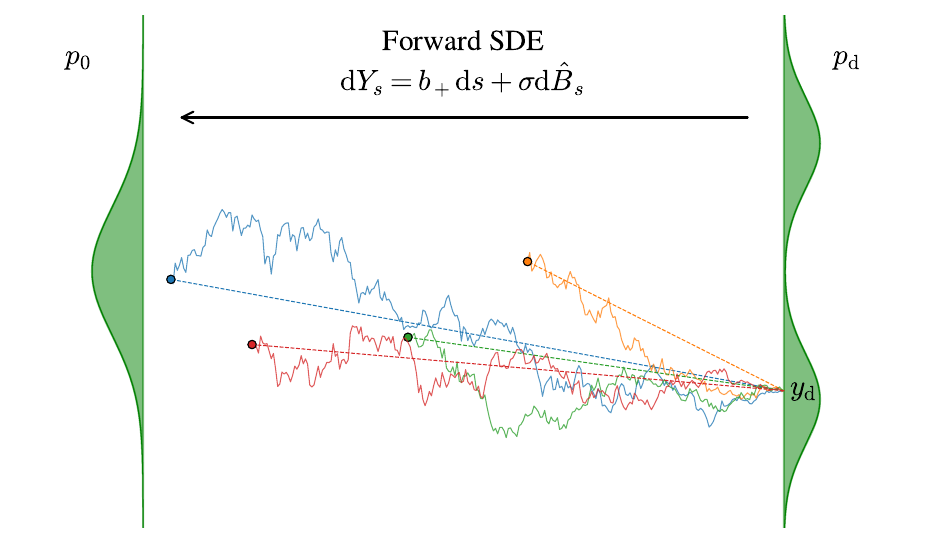}
    \caption{\label{fig:PathIntegralStochasticVsTransition} The Monte Carlo estimate of the path integral, \cref{eq:PathIntegralMC}, propagates $y_\d \sim \finP$ to several random instants of time by sampling the transition kernel $\P(y_\s, \s|y_\d,0)$. This allows us to reach the encircled points in one large jump (the dashed lines), thereby avoiding a full simulation of the stochastic trajectories. The accuracy of the MC estimate will improve if we use a larger number of such `throws'. See \cref{sec:FiniteNumberEffects}.}
\end{figure}

In a diffusion model we choose the forward process such that $\P(y_\s,\s | \finy, 0)$ is a Gaussian. This allows us to estimate \cref{eq:PathIntegralTransition} by Monte Carlo,
\begin{equation}
    \E_{\K}[\O[y(\s)]] =
        \T \, \E_{\finy \sim \finP}
        \E_{\s \sim \mathcal{U}(0,\T)} \E_{y_\s \sim \P(y_\s,\s|\finy, 0)} [f(y_\s,\s)] .
    \label{eq:PathIntegralMC}
\end{equation}
The key advantage is that we can evaluate the path integral \textit{without simulating the full stochastic evolution of each $\finy$}---using the Gaussian kernel we can propagate $\finy$ to any instant of time in a single step (see \cref{fig:PathIntegralStochasticVsTransition}). Furthermore, we can improve the MC estimate at no extra runtime cost by evolving each $\finy$ to several random points of time. This is why $\E_\s$ is nested inside $\E_{y_\d}$ in \cref{eq:PathIntegralMC}. We discuss this in greater detail in \cref{sec:FiniteNumberEffects}.

Diffusion models much easier to train compared to vanilla continuous normalizing flows (CNFs) \cite{Chen18,Grathwohl19}, because the training objective is a path integral that can be estimated efficiently by \cref{eq:PathIntegralMC}. In particular, this approach obviates the need to solve ODEs during training, replacing it with a simple average that can be computed efficiently. The path integral approach confers the same advantage on density estimation, as explained below.

\subsection{Density Estimation}
\label{sec:DensityEstimation}

Diffusion models can be formulated as a stochastic control problem \cite{Pavon89,Huang21}. Such an analysis leads to the following lower bound on the log likelihood of an individual datum $x$ \cite{Premkumar24a},
\begin{equation}
    \log \P_u(x,0) \geq
        - \E_{Y_{\s}}
        \left[
            \int_{0}^{T} \d \s \left(
                \frac{1}{2 \g^2} \left\| \bplus - u \right\|^2 - \nabla \cdot u
            \right)
            - \log \P_u(Y_\T, 0) \bigg| Y_0 = x
        \right] .
    \label{eq:LikelihoodBound}
\end{equation}
%
The bound is saturated if $u = \bminus$ (see App.\ B.1 of \cite{Premkumar24a}), in which case $\P_u(\cdot, \T) = \P_{\d}(\cdot)$. The conditional expectation value on the r.h.s.\ can be understood as follows: release a large number of `particles' from $x$ at time $s=0$, let them evolve under \cref{eq:ForwardSDE}, and calculate the average over the trajectories of each individual particle. At a glance this expression suffers from the same problems (\cref{pt:Autodiff,pt:Sequential}) as the ODE approach---we need to simulate stochastic trajectories and evaluate an integral that involves the gradient of a neural network, $\nabla \cdot u$. Worse still, we need to do compute the average over several such trajectories to evaluate the r.h.s.\ for a single $x$.

We can get around these issues by noting that, for a vector valued function $h(y_{\s},\s)$
\begin{equation}
    \begin{aligned}
        \E_{Y_{\s}} \left[ \nabla \cdot h(Y_{\s}, \s) | Y_0 = x \right]
            &= - \int \d y_s \, h(y_{\s}, \s) \cdot \nabla \P(y_{\s}, \s | x, 0) \\
            &= - \E_{Y_{\s}} \left[ h(Y_{\s}, \s) \cdot \nabla \log \P(Y_{\s}, \s | x, 0) | Y_0 = x \right] ,
        \label{eq:IBPonTransitionProbability}
    \end{aligned}
\end{equation}
where we have used integration by parts in the first step and assumed that the product $hp$ vanishes at the spatial boundaries. Using \cref{eq:IBPonTransitionProbability} in \cref{eq:LikelihoodBound} we obtain
\begin{align}
    \log \, &\P_u(x,0) \geq \label{eq:LikelihoodBoundEfficient} \\
        &- \E_{Y_{\s}}
        \left[
            \int_{0}^{T} \d \s \left(
                \frac{1}{2 \g^2} \left\| \bplus - u \right\|^2
                + u \cdot \nabla \log \P(Y_{\s}, \s|x, 0)
            \right)
            - \log \P_u(Y_\T, 0) \bigg| Y_0 = x
        \right] . \nn
\end{align}
By transferring the gradient operator from $u$ we avoid the need to take derivatives of the neural network; since the transition probability is a Gaussian the gradients of their log are easy to calculate. The r.h.s.\ is a path integral, so we can estimate it efficiently as shown in \cref{eq:PathIntegralMC}:
%
    \begin{equation}
        \begin{aligned}
            \log \, \P_u(x,0) \geq 
            &\E_{y_{T} \sim \P(y_{T}, T|x, 0)} \left[ \log \P_u(y_T) \right] \\[0.5em]
            &- \E_{\s \sim \mathcal{U}(0,T)} \E_{y_{\s} \sim \P(y_{\s}, \s|x, 0)}
            \left[
                \frac{1}{2 \g^2} \left\| \bplus - u \right\|^2
                + u \cdot \nabla \log \P(Y_{\s}, \s|x, 0)
            \right] .
        \end{aligned}
        \label{eq:MCEstimate}
    \end{equation}
%
This is the central result of this paper.
The path integral approach addresses all three issues mentioned in \cref{sec:ProbabilityFlowODE}:
\begin{enumerate}
    \item We do not have to simulate the SDE \cref{eq:ForwardSDE} to compute \cref{eq:MCEstimate}; the transition probability allows us to project the initial value $x$ to an arbitrary $\s$ in a single step. Since the time $\s$ is drawn from a uniform distribution over $[0,\T]$, the calculation can also be sped up by suitable vectorization.
    \item The path integral is agnostic to the value of $x$, or the drift and diffusion coefficients, so it computes $\log \P$ for all samples in nearly the same time. See \cref{fig:Fast_ODEvsPathIntegral}.
    \item We do not require derivatives of the neural network; we moved that derivative to the transition probability in \cref{eq:IBPonTransitionProbability}. The latter has a simple analytic expression.
\end{enumerate}

Both the ODE and path integral approaches can be vectorized to compute the log densities for multiple samples concurrently. However, the ODE method incurs an iterative overhead since the time integral must be calculated sequentially. On the other hand, the path integral estimates the solution by computing a Monte Carlo approximation at several points simultaneously, making it easier to scale, and more flexible in handling irregularities. These properties are manifest in the experiments in \cref{sec:ODEvsPathIntegral}.

\section{Experiments}
\label{sec:Experiments}

In order to understand the accuracy of our diffusion density estimator, we primarily work with lower-dimensional Gaussian mixture distributions for $\P_{\d}$. We sample these distributions to obtain $\{ \finy \}$, which are used to train a diffusion model. The neural network is a simple MLP with additional Gaussian random feature layers for embedding $x$ and $t$. The structure of the network was kept fixed in all experiments, and we used the ELBO loss from \cite{Ho20} to train the models. This is the same setup we used in \cite{Premkumar24a}. More details are given there.

In this work we are primarily interested in the path integral approach to density estimation, so the majority of our experiments are based on \cref{eq:MCEstimate}. Since we know the actual value of $\log \finP$, we can approximate the KL between the data and reconstructed distributions by computing log density on a large number of samples,
\begin{equation}
    \D_{KL} \left( \finP(\cdot) \| \P_u(\cdot, \T) \right)
        \approx \frac{1}{N_{\rm test}} \sum_{k=1}^{N_{\rm test}} \log \finP(x^{(k)}) - \log \P_u(x^{(k)}, \T) .
    \label{eq:KLbyMC}
\end{equation}
A few points to note: First, we substitute the bound from \cref{eq:MCEstimate} in place of $\log \P_u$ in these expressions. Therefore, what we are actually computing is the l.h.s.\ from \cref{eq:BoundOnKLGeneral}, the upper bound on the KL. Henceforth all mention of KL divergence is actually referring to its maximum value. Second, due to imperfections in learning the result of evaluating \cref{eq:MCEstimate} can be quite noisy. In fact it can be larger than $\log \finP(x^{(k)})$ in some cases (see \cref{fig:Fast_Likelihoods}). To address these fluctuations we evaluated the KL at $N_{\rm test} = 10^4$ points in each experiment, and repeated each experiment with 8 different seed values to initialize the random number generators and the neural network parameters. Under these conditions \cref{eq:KLbyMC} consistently gives a non-negative answer. We use this KL divergence as a gauge of the model performance in all our experiments.

\subsection{Finite Number Effects}
\label{sec:FiniteNumberEffects}

\begin{figure}
    \centering
    \caption{\label{fig:TrainingParameters} \textbf{The effect of training parameters on model performance.}}
    \bigskip
    
    \begin{subfigure}{\linewidth}
        \centering
        \includegraphics[width=\linewidth]{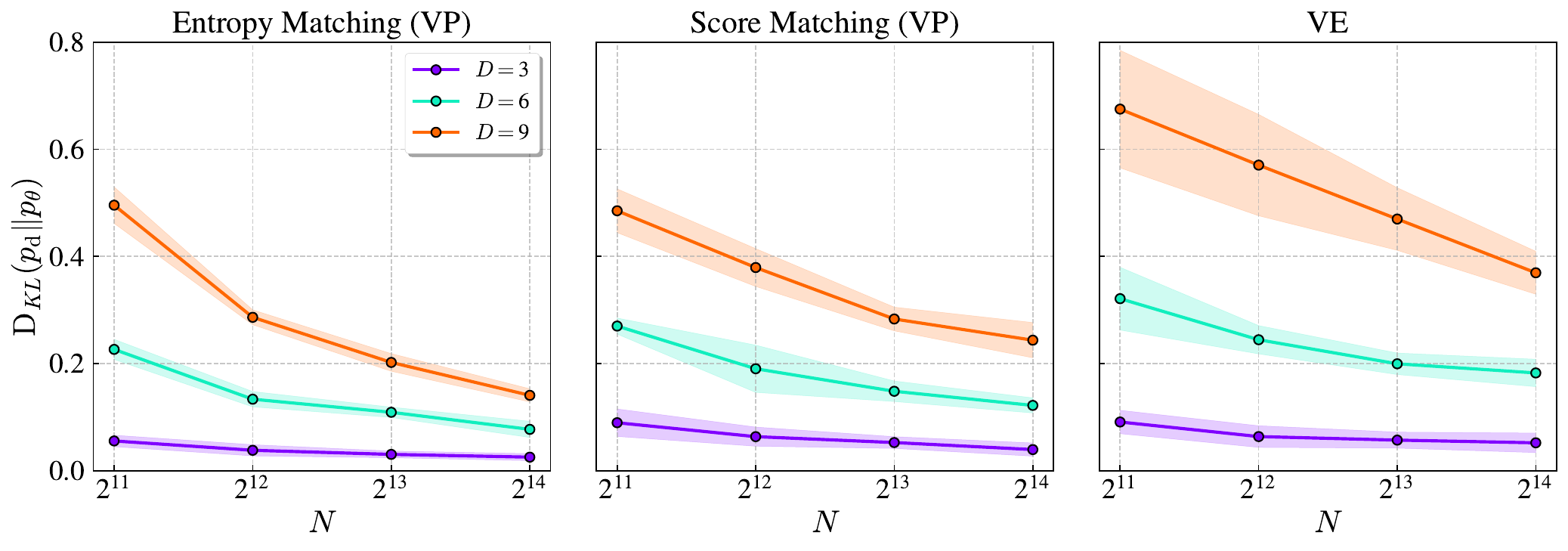}
        \subcaption{\label{fig:Fast_KLvN} KL vs.\ number of samples. Trained with $n_t = 10$ throws over $n_{\rm ep}=200$ epochs.}
    \end{subfigure}
    \par\bigskip 
    \begin{subfigure}{\linewidth}
        \centering
        \includegraphics[width=\linewidth]{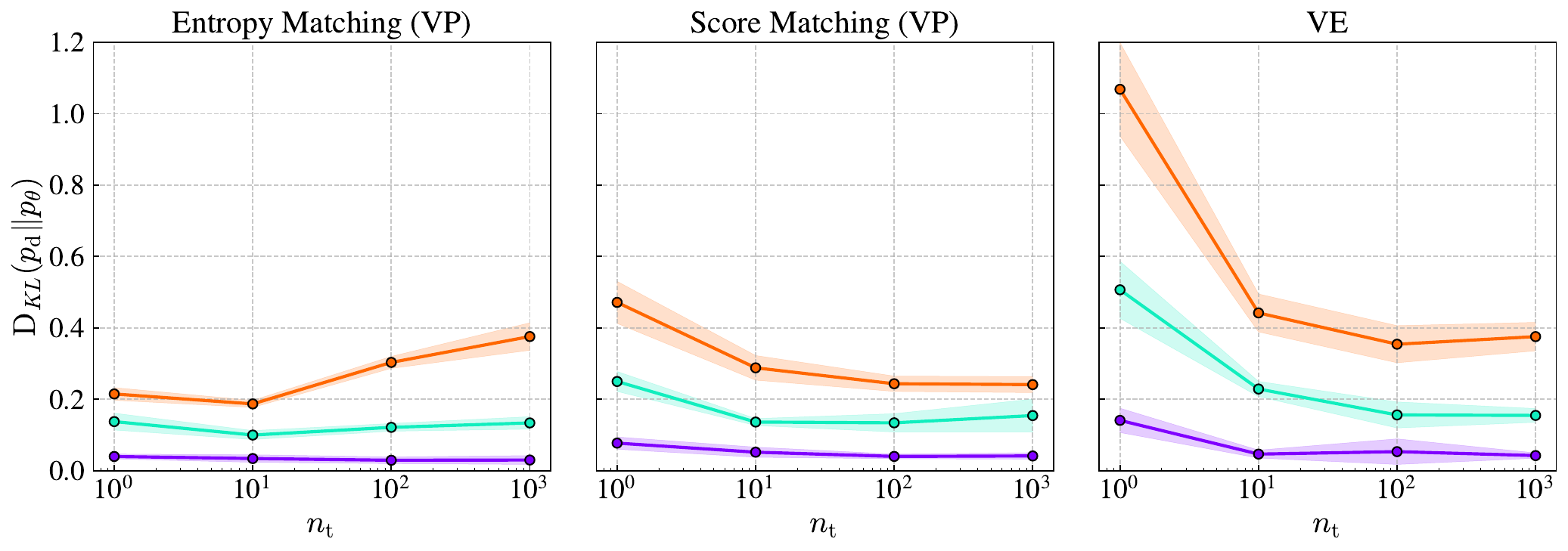}
        \subcaption{\label{fig:Fast_KLvNsteps} KL vs.\ number of throws. Trained on $N=8192$ samples over $n_{\rm ep}=200$ epochs.}
    \end{subfigure}
    \par\bigskip 
    \begin{subfigure}{\linewidth}
        \centering
        \includegraphics[width=\linewidth]{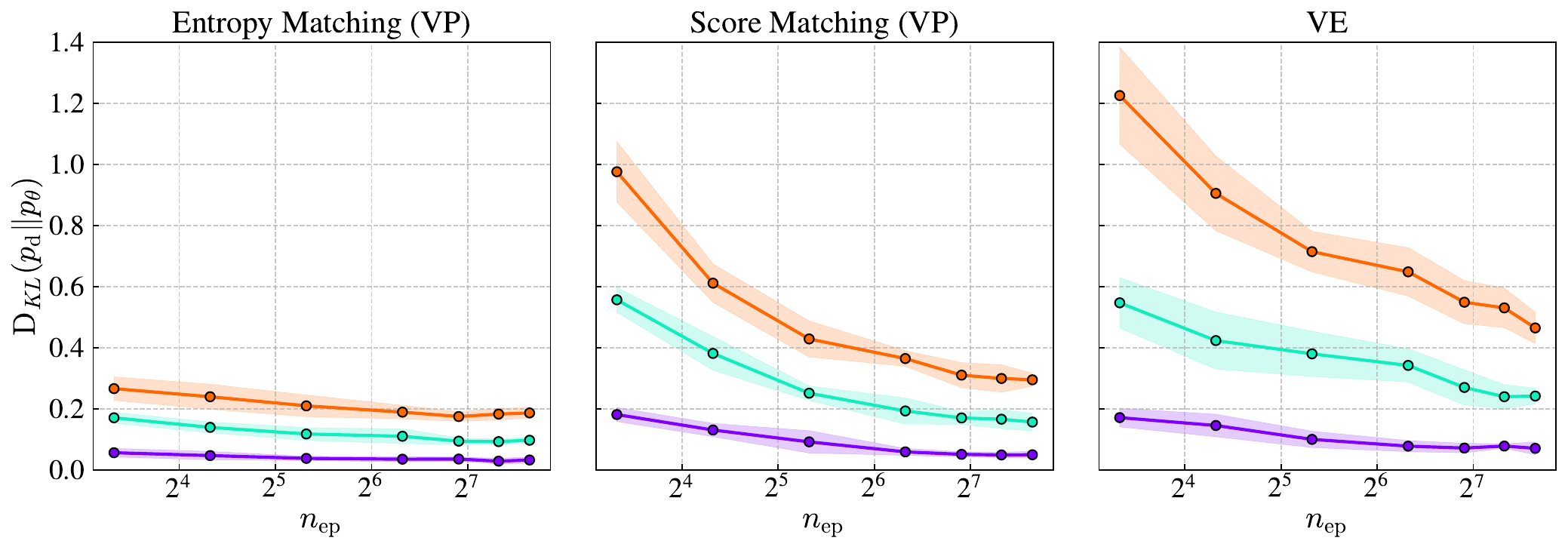}
        \subcaption{\label{fig:Fast_KLvEpochs} KL vs.\ number of training epochs. Trained on $N=8192$ samples with $n_t = 10$.}
    \end{subfigure}
\end{figure}

Both the loss function of a diffusion model (cf.\ \cref{eq:DenoisingSM,eq:DenoisingEM}), as well as the lower bound on the log density, \cref{eq:LikelihoodBoundEfficient}, are path averages of the form discussed in \cref{sec:PathAverages}. In reality, this bound is not saturated since no neural network can reproduce $u = \bminus$ perfectly; the loss function is never zero. To a large extent, these limitations are a manifestation of the Law of Large Numbers; both the loss and the log density bound are averages that approach the true value when evaluated over a great many sample points. However, in practice we only have a finite amount of data and compute, so we are limited in how close we can get to the actual answer. The key drivers of model performance are the number of training samples ($N$), the number of throws ($n_{\rm t}$) per sample, and the number of training epochs ($n_{\rm ep}$). In every training epoch, each of the $N$ samples is projected forward to $n_{\rm t}$ random points of time in the interval $(0, \T]$, with the $n_{\rm t}$ instants differing from one epoch to another. Therefore, the loss function is evaluated over a total of $N \times n_{\rm t} \times n_{\rm ep}$ points. The effect of varying each of these parameters is shown in \cref{fig:TrainingParameters}. The main characteristics of these plots are discussed below.

\par\medskip
\noindent\textbf{Number of training samples} ($N$): A larger value of $N$ helps us reconstitute the underlying distribution more faithfully \cite{Bialek96,Premkumar23}. This is unsurprising, since the diffusion model converts the prior to the target density using information gathered from the samples $\{ \finy \}$. At the same time, the training data becomes sparser at higher $\Dim$, so more samples are required to achieve the same model performance as the dimensionality increases. See \cref{fig:Fast_KLvN}.

\par
\begin{figure}
    \centering
    \caption{\label{fig:ThrowsAndEpochs} \textbf{Interplay between number of throws and training epochs.} $N=8192$.}
    
    \begin{subfigure}{\linewidth}
        \centering
        \includegraphics[width=0.955\linewidth]{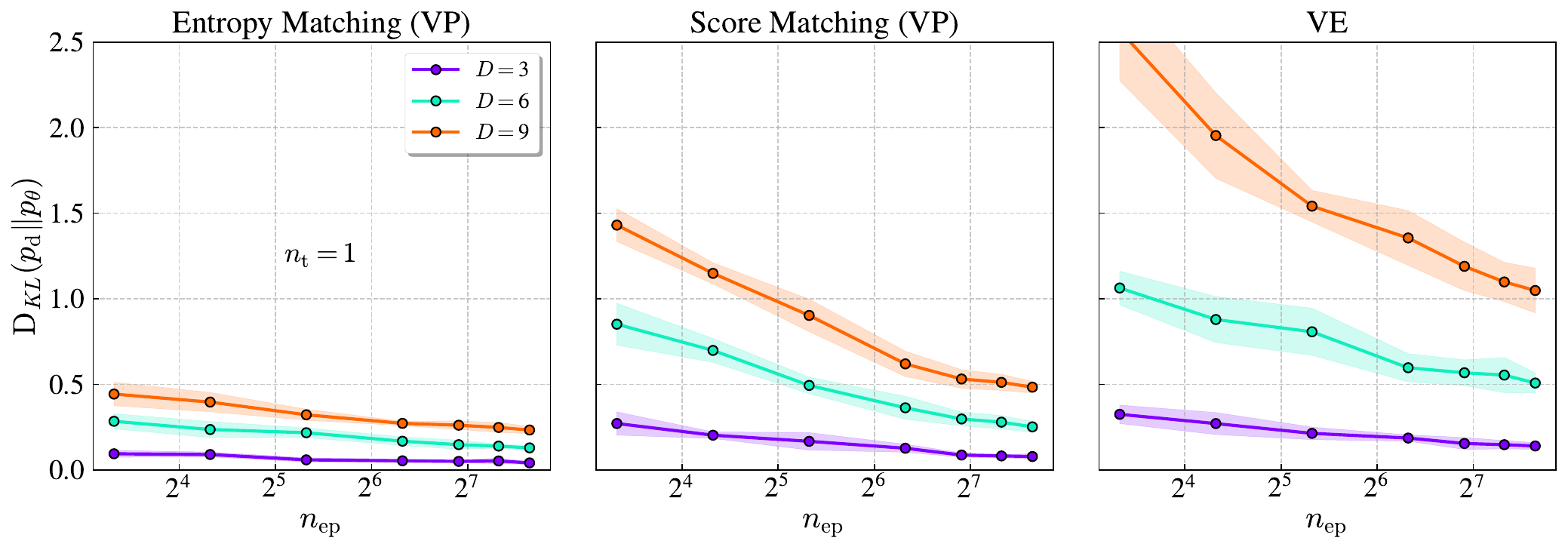}
    \end{subfigure}
    \begin{subfigure}{\linewidth}
        \centering
        \includegraphics[width=0.955\linewidth]{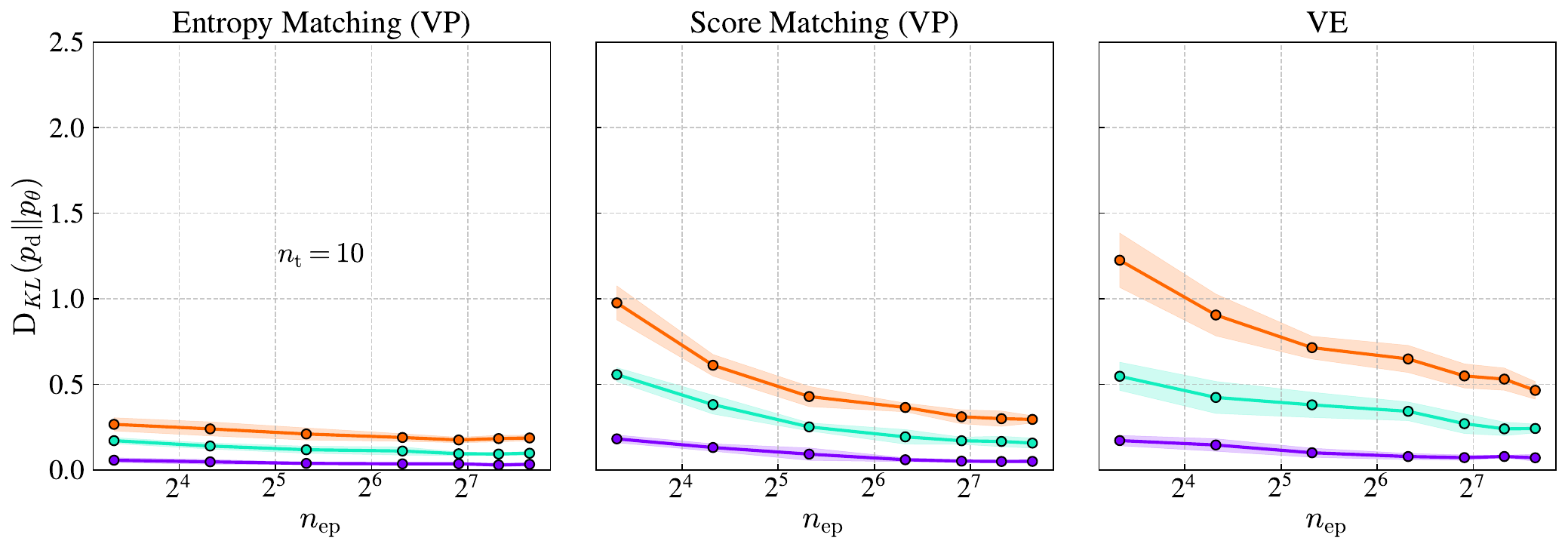}
    \end{subfigure}
    \begin{subfigure}{\linewidth}
        \centering
        \includegraphics[width=0.955\linewidth]{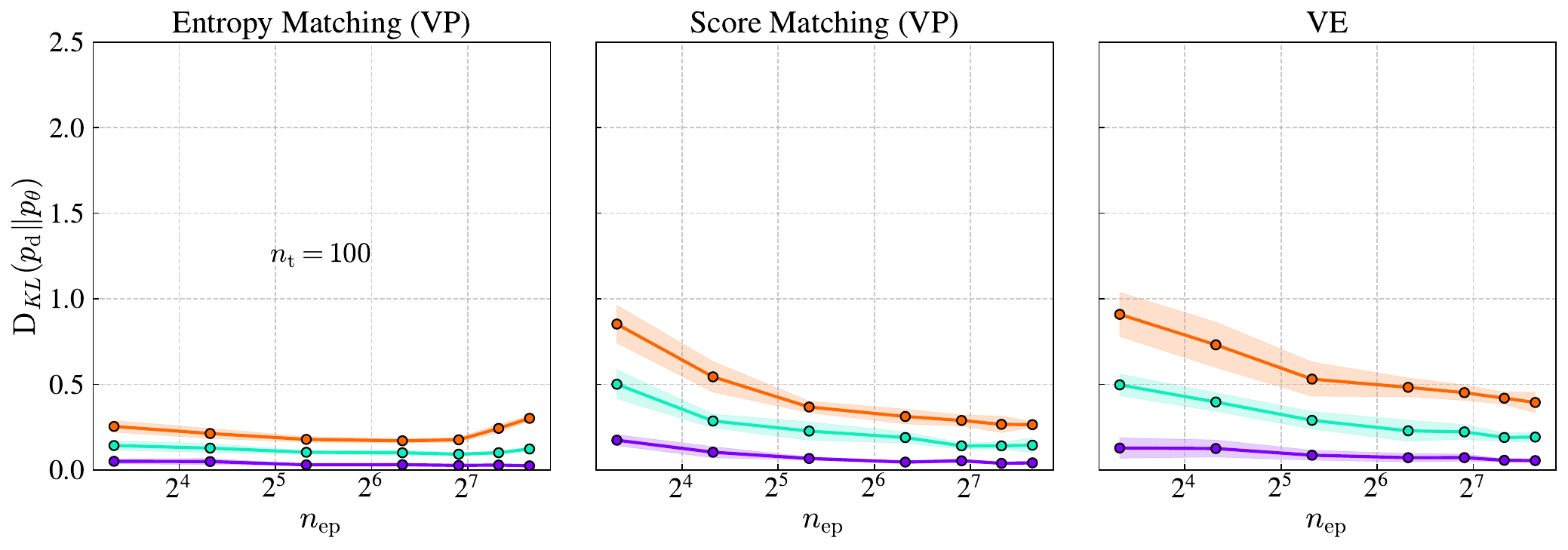}
    \end{subfigure}
    \begin{subfigure}{\linewidth}
        \centering
        \includegraphics[width=0.955\linewidth]{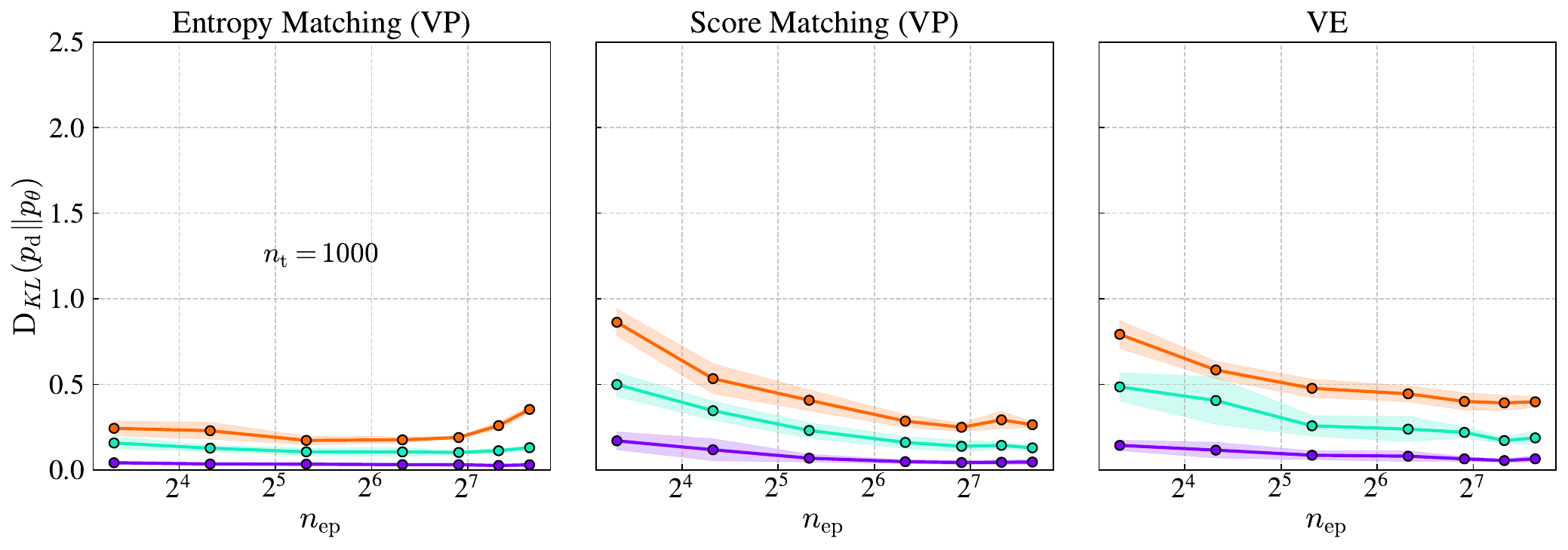}
    \end{subfigure}
\end{figure}

\medskip
\noindent\textbf{Number of throws per sample} ($n_{\rm t}$):
Each $y_\d$ is thrown $n_{\rm t}$ times in our training process (for example, $n_{\rm t} = 4$ in \cref{fig:PathIntegralStochasticVsTransition}). In the score matching VP and VE models, there is a notable improvement in the KL as $n_{\rm t}$ is increased, as shown in \cref{fig:Fast_KLvNsteps}. But the opposite appears to be true for entropy matching: these models learn better with fewer throws, but worsen at larger $n_{\rm t}$. The number of training epochs influences this behavior. Further experiments show that, \textit{for an entropy matching model, fewer throws and shorter training durations sometimes yield better results} than using more throws and longer training. See the $\Dim=9$ plots in \cref{fig:ThrowsAndEpochs}, particularly for $n_{\rm t}=100,1000$.

\par\medskip
\noindent\textbf{Number of training epochs} ($n_{\rm ep}$): Once again, the score matching VP and VE models improve in performance if training is carried out for longer. Remarkably, the entropy matching model learns much more rapidly, approaching near-terminal performance in less than 10 epochs. 

Taken together, these observations provide strong motivation for further research into entropy matching models. If the properties of this model generalize to real-world datasets, it could significantly reduce the memory, time, and energy costs of training diffusion models. We leave this investigation for future work. In the interim, we recall the following facts about entropy matching, which could be potential clues of their efficiency: (1) entropy matching models deliver to the network \textit{only} the information needed to convert the prior to the target distribution, compared to score matching wherein the network also needs to retain information to counter the drift term \cite{Premkumar24a}, and (2) the information is localized to a small interval within $(0,\T]$, as evident from the entropy production profile.


It is also worth noting that the VE process consistently underperforms in our benchmarks, but also gains the most from larger values of $N$, $n_{\rm t}$, and $n_{\rm ep}$. The underlying reason is that, in the absence of a drift term, the forward process spreads the distribution over a larger volume, diluting $\nabla \log \P$ in the process. There are two consequences: (1) the score values are too small to estimate accurately in most regions at late $\s$, and (2) a larger and finer grid of points is needed to evaluate \cref{eq:PathIntegralMC} accurately. On the other hand, the VP process transforms the data into a Gaussian ball of roughly unit radius, therefore the volume effects that plague VE models are subdued to a great extent in VP. However, the volume of this ball grows with $\Dim$, which is why the KL increases with dimensionality in all plots in \cref{fig:TrainingParameters,fig:ThrowsAndEpochs}. Notably, the deterioration in performance at larger $\Dim$ is most gradual in entropy matching.

Finally, we point out that the bound in \cref{eq:MCEstimate} is evaluated over a single test point, $x$, unlike the loss function which is evaluated over all $N$ samples in $\{ \finy \}$. Therefore, the quality of that estimate depends solely on the number throws of $x$, which was set to $n_{\rm t} = 10^5$ in all cases.




\subsection{ODE vs.\ Path Integral}
\label{sec:ODEvsPathIntegral}

\begin{figure}
    \centering
    \includegraphics[width=\linewidth]{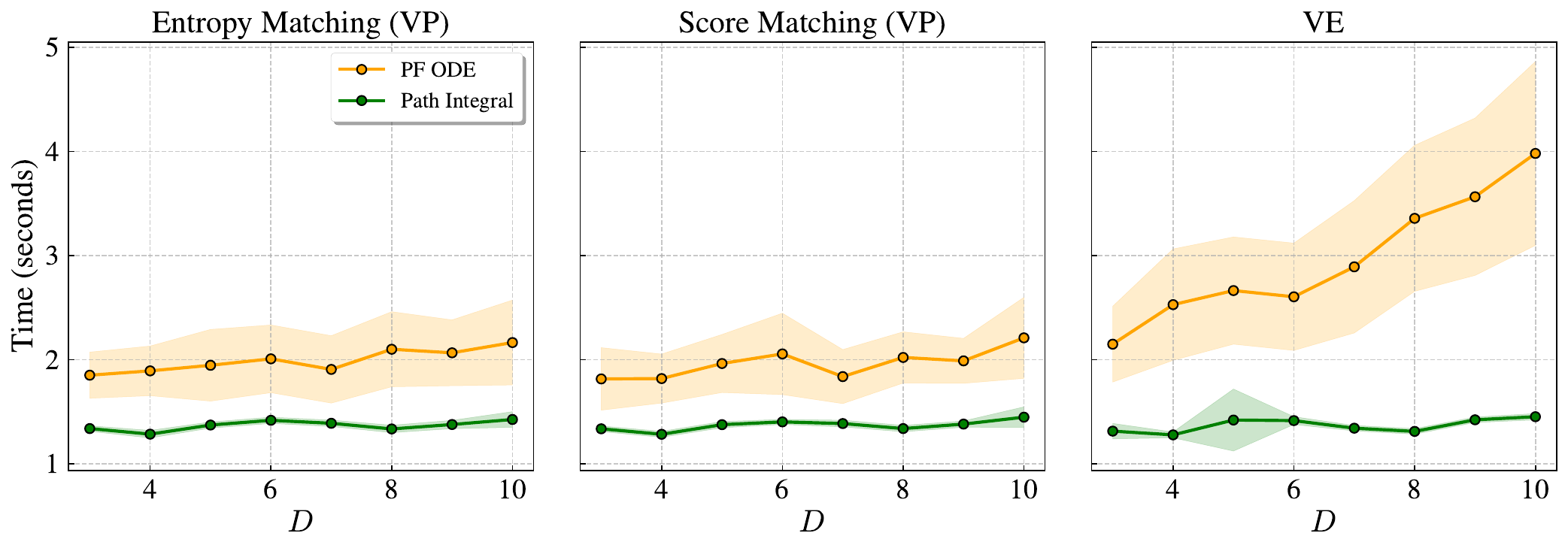}
    \caption{\label{fig:Fast_ODEvsPathIntegral} Time per sample to compute $\log p$ with the probability flow ODE and path integral approaches, with the same accuracy. The horizontal axis is dimensionality. All models were trained with $N=8192$, $n_{\rm t} = 10$, and $n_{\rm ep} = 200$.}
\end{figure}

Given the same diffusion model, the Probability Flow ODE (PF ODE) and path integral approaches can estimate log density to comparable accuracy. Each approach introduces its own source of error: in the ODE case it is the trace estimator and the black-box solver (see \cref{sec:TraceEstimator}), and for path integrals it is the error in estimating \cref{eq:MCEstimate} with a finite number of throws. We found that $n_\ep = 1000$ normal random vectors in the trace estimator, and $n_{\rm t} = 10^5$ throws in the path integral gave similar results. We used both methods to evaluate the log density over the same 100 samples, one at a time, and profiled the runtime. The results are shown in \cref{fig:Fast_ODEvsPathIntegral}.
\AP{A more careful experiment would use different $n_{\rm t}$ for the different models.}

Unsurprisingly, the ODE is slower than the path integral, for reasons discussed in \cref{sec:DensityEstimation}. The larger variance in time for the ODE is due to adaptive step sizes used by the RK45 solver. It is also the reason why the ODE approach is slower at larger $\Dim$ in the VE case; the smaller values of $\nabla \log \P$ at early $\t$ cause the solver to take smaller steps to reduce error. The path integral approach takes nearly the same time for all samples since it is a simple average.

\section{Conclusions}
\label{sec:Conclusions}

We have shown that diffusion models can be used as neural density estimators, and introduced a new approach to extract log densities from these models. Through our experiments, we studied the performance of different models under a range of training parameters. We found that the entropy matching model trains more quickly and with a smaller memory footprint, while delivering results on par with, and often surpassing, those of the score matching VP and VE models. These observations strongly motivate further research into entropy matching models, particularly for images and other higher-dimensional distributions, where their efficiency and performance may offer substantial benefits.

We used samples from simple Gaussian mixtures in low dimensions to train our diffusion models, although a normalizing flow method like MAF would make a better density estimator for such distributions \cite{nflows}. Our experiments serve as proof of concept, paving the way for future research into potential applications of diffusion models for log density computation. As density estimators , these models could prove valuable in simulation-based likelihood-free inference \cite{Fan12} or for learning proposal distributions in Monte Carlo methods. Furthermore, leveraging diffusion models as zero-shot classifiers presents another exciting avenue for exploration \cite{Li23}.

\subsection*{Acknowledgements}

We thank Lorenzo Orecchia, William Cottrell, Antares Chen, and Di Luo for valuable discussions. The author was supported in part by the Kavli Institute for Cosmological Physics at the University of Chicago through an endowment from the Kavli Foundation. Computational resources for this project were made available through the AI+Science research funding from the Data Science Institute at the University of Chicago.

\newpage
\appendix

\section{Parameterization of the control}
\label{sec:Parameterization}

The control $u$ from \cref{sec:ReverseDiffusion} can be expressed as $u = u_{a} + u_\th$, where $u_{a}$ has a closed-form expression and $u_\th$ is the output of a neural network. The choice of how we split $u$ into these components dictates the information retained by the neural network \cite{Premkumar24a}. We consider two parameterizations of the control below.

\subsection{Score Matching}

If we choose $u = \bplus - \g^2 {\bm s}_\th$, where ${\bm s}_\th$ is the output of a neural network, the upper bound from \cref{eq:BoundOnKLGeneral} becomes \cite{Song21}
\begin{equation}
    \int_{0}^{T} {\rm d} \s \, \g^2(\s) \E_{Y_{\s}} \left[ \frac{1}{2} \big\| {\bm s}_\th(Y_{\s}, {\s}) - \nabla \log \K(Y_{\s}, \s) \big\|^2 \right] ,
    \label{eq:ExplicitScoreMatching}
\end{equation}
Here $\E_{Y_{\s}} \left[ g(Y_{\s}, \s) \right] \deq \int {\rm d} y_\s \, \K(y_\s,\s) g(y_\s,\s) \equiv \E_{\K} \left[ g(Y_{\s}, \s) \right]$ is the expectation value with respect to distribution $\K$. We do not have access to $\K$ or $\nabla \log \K$, so we cannot compute \cref{eq:ExplicitScoreMatching}. However, it can be shown that \cref{eq:ExplicitScoreMatching} is equivalent to the denoising score-matching objective \cite{Vincent11}
\begin{equation}
    \int_{0}^{T} {\rm d} \s \, \g^2(\s) \E_{Y_{\s}, Y_0} \left[ \frac{1}{2} \big\| {\bm s}_\th(Y_{\s}, {\s}) - \nabla \log \P(Y_{\s}, \s | Y_0, 0) \big\|^2 \right] . \label{eq:DenoisingSM}
\end{equation}
If the drift term $\bplus(Y,\s)$ is zero or linear in $Y$, the transition probability $\P(Y_{\s}, \s | Y_0, 0)$ is a Gaussian, and the gradient of its log has a closed form expression. Furthermore, given the samples $\{ \finy \}$ of $\P_{\d}$ we can form a Monte Carlo estimate of \cref{eq:DenoisingSM}, as explained in \cref{sec:PathIntegrals}. Once the network is trained, we can generate a sample $x(T) \sim \P_{\d}$ by drawing a sample $x(0) \sim \P_0$ and evolving it with
%
\begin{equation}
    \d X_{\t} = - (\bplus - \g^2 {\bm s}_\th ) (X,T-\t)  \d \t + \g(T-\t) \d \hat{\B}_{\t} . \label{eq:SMReverseSDE}
\end{equation}

\subsection{Entropy Matching}

With the choice $u = - \bplus - \g^2 \e_\th$, the upper bound in \cref{eq:BoundOnKLGeneral} becomes
\begin{equation}
    \int_{0}^{\T} \d \s \, \g^2(\s) \E_{Y_\s} \left[
        \frac{1}{2} \left\lVert \frac{2 b_{+}(Y_\s,\s)}{\g^2(\s)} - \nabla \log \K(Y_\s,\s) + \e_\th(Y_\s,\s) \right\rVert^2
    \right]
\end{equation}
It is beneficial to parameterize the control this way because we can assess the amount of information delivered to the neural network during training \cite{Premkumar24a}. It has been empirically observed that the density estimates worsen when the network has to retain more information. Following the same steps that led to \cref{eq:DenoisingSM}, we obtain the denoising entropy matching objective
\begin{equation}
    \int_{0}^{\T} \d \s \, \g^2(\s) \E_{Y_\s,Y_0} \left[
        \frac{1}{2} \left\lVert \frac{2 b_{+}(Y_\s,\s)}{\g^2(\s)} - \nabla \log \P(Y_\s,\s|Y_0,0) + \e_\th(Y_\s,\s) \right\rVert^2
    \right] .
    \label{eq:DenoisingEM}
\end{equation}
The generative SDE is now
\begin{equation}
    \d X_{\t} = (\bplus + \g^2 \e_\th ) (X,T-\t)  \d \t + \g(T-\t) \d \hat{\B}_{\t} . \label{eq:EMReverseSDE}
\end{equation}
Score matching and entropy matching become identical when the drift term in the forward process is set to zero, $\bplus = 0$.

\section{Trace Estimator}
\label{sec:TraceEstimator}

We review the Hutchinson trace estimator and provide practical tips for vectorizing its use in computing neural network derivatives.

\begin{figure}
    \centering
    \includegraphics[width=\linewidth]{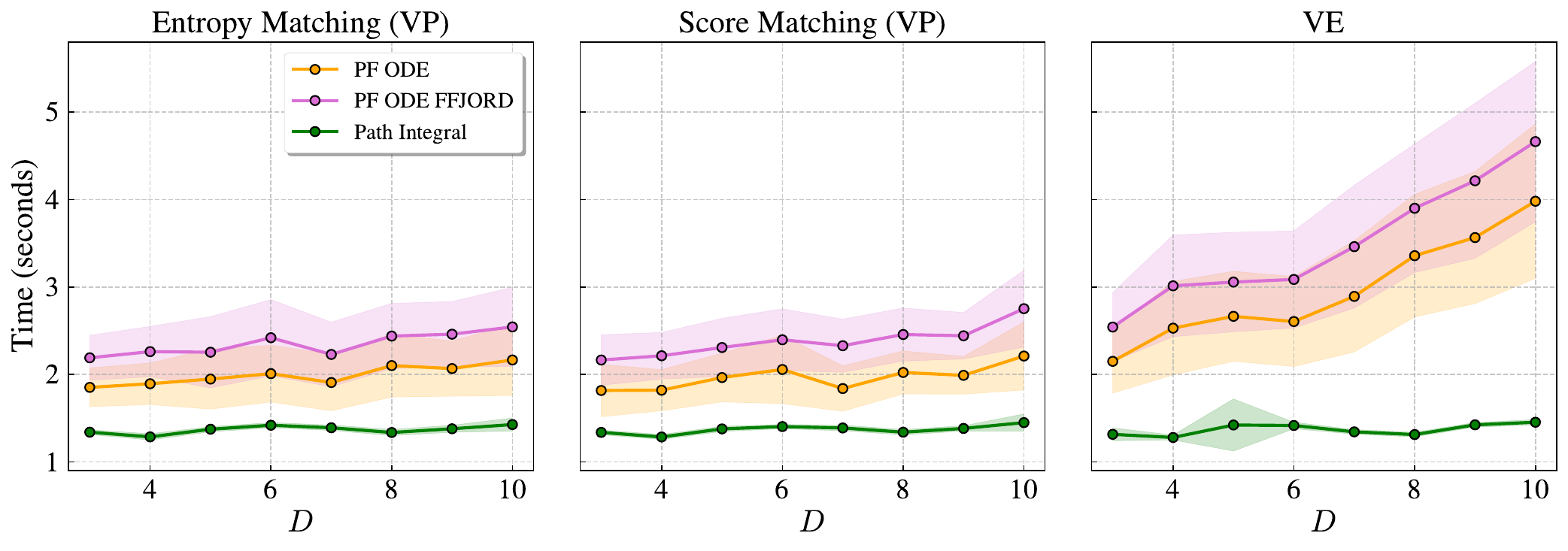}
    \caption{\label{fig:Fast_ODEvsPathIntegralDelayed} Time per sample taken by the ODE approach when the average over the random $\ep$ vectors are computed only \textit{after} evaluating the time integral. This is the `PF ODE FFJORD' plot, shown together with the data from \cref{fig:Fast_ODEvsPathIntegral}.}
\end{figure}

Consider \cref{eq:ODESystem}. To solve for the $\log$ term we need to compute $\nabla \cdot f$, where $f$ involves a neural network that approximates $\nabla \log \P$ (cf.\ \cref{eq:ReverseLiouville}). For example, in a VE model, the log density at $\T$ is given by \cite{Song21}
\begin{equation}
    \log \finP(x_\T) = \log \inP(x_0) - \frac{1}{2} \int_{0}^{\T} \d \t \, \bar{\g}^{2(\T-\t)} \nabla \cdot {\bm s}_\th (x,\t) ,
    \label{eq:ScoreMatchingLogP}
\end{equation}
where $\bar{\g}$ is a constant, and ${\bm s}_\th$ is a deep neural network. The divergence of ${\bm s}_\th$ is simply the trace of its Jacobian, $J_{{\bm s}_\th}$. The latter can be approximated by the \textit{Skilling-Hutchinson trace estimator} \cite{Hutchinson90,Adams18}. We can estimate the trace of a matrix $A$ by multiplying it with random vectors and taking the average. In the Einstein notation,
\begin{equation}
    {\rm tr} (A)
        = A_{ij} \delta_{ji}
        \approx A_{ij} \E \left[ \ep_j \ep_i \right]
        = \E \left[ (A_{ij} \ep_j) \ep_i \right]
        = \E \left[ \ep_i A_{ij} \ep_j \right]
        = \E \left[ \ep^{\top} \cdot A \cdot \ep \right] ,
\end{equation}
Here $\ep$'s are random vectors sampled from a distribution with mean 0 and unit variance along each direction, for e.g.\ Rademacher or Gaussian (we denote it by $\mathcal{P}(0, \mathbf{1})$). In the second step we have used $\E \left[ \ep \ep^{\top} \right] \approx \mathds{1}$. We can use the estimator to compute the divergence in \cref{eq:ScoreMatchingLogP},
\begin{equation}
    \nabla \cdot {\bm s}_\th = {\rm tr}(J_{{\bm s}_\th})
        \approx \E_{\ep \sim \mathcal{P}(0, \mathds{1})} \left[ \ep^{\top} \cdot J_{{\bm s}_{\bm \theta}} \cdot \ep \right] .
    \label{eq:TraceEstimator}
\end{equation}
Computing $\nabla \cdot {\bm s}_\th$ costs $\O(\Dim^2)$, whereas $\ep^{\top} \cdot J_{{\bm s}_\th}$ can be computed for nearly the same cost as evaluating ${\bm s}_\th$ using reverse-mode automatic differentiation. That is, $\ep^{\top} \cdot J_{{\bm s}_\th} = (\nabla (\ep^{\top} \cdot {\bm s}_\th))^{\top}$, and so
\begin{equation}
    \nabla \cdot {\bm s}_\th \approx \E_{\ep \sim \mathcal{P}(0, \mathds{1})} \left[ (\nabla (\ep^{\top} \cdot {\bm s}_\th))^{\top} \cdot \ep \right] .
    \label{eq:GradWithTraceEstimator}
\end{equation}
\cref{eq:GradWithTraceEstimator} is then
\begin{equation}
    \log \finP(x_\T) = \log \inP(x_0) - \frac{1}{2} \int_{0}^{\T} \d \t \, \bar{\g}^{2(\T-\t)} \E_{\ep \sim \mathcal{P}(0, \mathds{1})} \left[ (\nabla (\ep^{\top} \cdot {\bm s}_\th))^{\top} \cdot \ep \right] ,
\end{equation}
The average can be moved outside the time integral, as done in FFJORD \cite{Grathwohl19}, but we observed empirically that this slowed down the solver without any notable improvement in accuracy (see \cref{fig:Fast_ODEvsPathIntegralDelayed}).

The expectation value can be computed efficiently by forming a sum over \textit{all} $n_\ep$ random vectors, and then taking the derivative. This is easier to understand in the Einstein notation. We start by writing $\nabla_x (\ep^{\top} \cdot {\bm s}_\th) \equiv \pd_i \ep_j {\bm s}_j$. Let $x$ be a batch of $n_\ep$ copies of the same sample, which means the indices $i,j$ run over $\Dim n_\ep$ values. We can split each of these indices into two. $i \to \mu a$ and $j \to \nu b$, where the $\mu, \nu$ indices run over $n_\ep$ values, and $a,b$ run over the $D$ coordinates. Then, $\pd_i \ep_j {\bm s}_j \equiv \pd_{\mu a} \ep_{\nu b} {\bm s}_{\nu b}$ can be understood as an $n_\ep \times \Dim$ matrix. We can contract this derivative with the matrix of all $\ep$'s,
\begin{equation}
    \ep_{\mu a} \pd_{\mu a} \ep_{\nu b} {\bm s}_{\nu b} = \sum_{\mu a} \sum_{\nu b} \ep_{\mu a} \pd_{\mu a} \ep_{\nu b} {\bm s}_{\nu b} .
\end{equation}
We have written out the sums explicitly for clarity. Since we are using the same sample, we can drop the batch indices from the derivative and the score function, $\pd_{\mu a} = \pd_{a}$ and ${\bm s}_{\nu b} = {\bm s}_{b}$. Dividing the sum by $n_\ep$ gives us the average in \cref{eq:GradWithTraceEstimator}:
\begin{align}
    \frac{1}{n_\ep} \sum_{\mu a} \sum_{\nu b} \ep_{\mu a} \pd_{a} \ep_{\nu b} {\bm s}_{b}
        &= \frac{1}{n_\ep} \sum_{a b} \pd_{a} {\bm s}_{b} \sum_{\mu \nu} \ep_{\mu a} \ep_{\nu b} \nn \\
        &= \frac{1}{n_\ep} \sum_{a b} \pd_{a} {\bm s}_{b}
            \left(
                \sum_{\mu=1}^{n_\ep} \ep_{\mu a} \sum_{\substack{\nu=1 \\ \nu \neq \mu}}^{n_\ep} \ep_{\nu b} +
                \sum_{\mu=1}^{n_\ep} \ep_{\mu a} \ep_{\mu b}
            \right) \nn \\
        &\approx \sum_{a b} \pd_{a} {\bm s}_{b} \left( \frac{1}{n_\ep}  \sum_{\mu=1}^{n_\ep} \ep_{\mu a} \ep_{\mu b} \right) \nn \\
        &\approx \sum_{a} \pd_{a} {\bm s}_{a}
        = \nabla \cdot {\bm s}_\th
\end{align}
In the second step used the fact that, for large enough $n_\ep$, $\sum_\nu \ep_{\nu b} \propto \E[\ep]_{b} \approx 0$, and in the last step $\frac{1}{n_\ep}  \sum_{\mu} \ep_{\mu a} \ep_{\mu b} = \E[\ep \ep^{\top}]_{ab} \approx \delta_{ab}$.

\clearpage
\phantomsection
\addcontentsline{toc}{section}{References}
\small
\bibliographystyle{unsrtnat}
\bibliography{fast}

\begin{thebibliography}{25}
\providecommand{\natexlab}[1]{#1}
\providecommand{\url}[1]{\texttt{#1}}
\expandafter\ifx\csname urlstyle\endcsname\relax
  \providecommand{\doi}[1]{doi: #1}\else
  \providecommand{\doi}{doi: \begingroup \urlstyle{rm}\Url}\fi

\bibitem[Bialek et~al.(1996)Bialek, Callan, and Strong]{Bialek96}
William Bialek, Curtis~G. Callan, and Steven~P. Strong.
\newblock Field theories for learning probability distributions.
\newblock \emph{Phys. Rev. Lett.}, 77:\penalty0 4693--4697, Dec 1996.
\newblock \doi{10.1103/PhysRevLett.77.4693}.
\newblock URL \url{https://link.aps.org/doi/10.1103/PhysRevLett.77.4693}.

\bibitem[Dinh et~al.(2017)Dinh, Sohl{-}Dickstein, and Bengio]{Dinh17}
Laurent Dinh, Jascha Sohl{-}Dickstein, and Samy Bengio.
\newblock Density estimation using real {NVP}.
\newblock In \emph{5th International Conference on Learning Representations, {ICLR} 2017, Toulon, France, April 24-26, 2017, Conference Track Proceedings}. OpenReview.net, 2017.
\newblock URL \url{https://openreview.net/forum?id=HkpbnH9lx}.

\bibitem[Papamakarios et~al.(2017)Papamakarios, Pavlakou, and Murray]{Papamakarios17}
George Papamakarios, Theo Pavlakou, and Iain Murray.
\newblock Masked autoregressive flow for density estimation.
\newblock In I.~Guyon, U.~Von Luxburg, S.~Bengio, H.~Wallach, R.~Fergus, S.~Vishwanathan, and R.~Garnett, editors, \emph{Advances in Neural Information Processing Systems}, volume~30. Curran Associates, Inc., 2017.
\newblock URL \url{https://proceedings.neurips.cc/paper_files/paper/2017/file/6c1da886822c67822bcf3679d04369fa-Paper.pdf}.

\bibitem[Grathwohl et~al.(2019)Grathwohl, Chen, Bettencourt, Sutskever, and Duvenaud]{Grathwohl19}
Will Grathwohl, Ricky T.~Q. Chen, Jesse Bettencourt, Ilya Sutskever, and David Duvenaud.
\newblock {FFJORD:} free-form continuous dynamics for scalable reversible generative models.
\newblock In \emph{7th International Conference on Learning Representations, {ICLR} 2019, New Orleans, LA, USA, May 6-9, 2019}. OpenReview.net, 2019.
\newblock URL \url{https://openreview.net/forum?id=rJxgknCcK7}.

\bibitem[Song et~al.(2021)Song, Sohl{-}Dickstein, Kingma, Kumar, Ermon, and Poole]{Song21}
Yang Song, Jascha Sohl{-}Dickstein, Diederik~P. Kingma, Abhishek Kumar, Stefano Ermon, and Ben Poole.
\newblock Score-based generative modeling through stochastic differential equations.
\newblock In \emph{9th International Conference on Learning Representations, {ICLR} 2021, Virtual Event, Austria, May 3-7, 2021}. OpenReview.net, 2021.
\newblock URL \url{https://openreview.net/forum?id=PxTIG12RRHS}.

\bibitem[Chen et~al.(2018)Chen, Rubanova, Bettencourt, and Duvenaud]{Chen18}
Tian~Qi Chen, Yulia Rubanova, Jesse Bettencourt, and David Duvenaud.
\newblock Neural ordinary differential equations.
\newblock In Samy Bengio, Hanna~M. Wallach, Hugo Larochelle, Kristen Grauman, Nicol{\`{o}} Cesa{-}Bianchi, and Roman Garnett, editors, \emph{Advances in Neural Information Processing Systems 31: Annual Conference on Neural Information Processing Systems 2018, NeurIPS 2018, December 3-8, 2018, Montr{\'{e}}al, Canada}, pages 6572--6583, 2018.
\newblock URL \url{https://proceedings.neurips.cc/paper/2018/hash/69386f6bb1dfed68692a24c8686939b9-Abstract.html}.

\bibitem[Nelson(1966)]{Nelson66}
Edward Nelson.
\newblock Derivation of the schr\"odinger equation from newtonian mechanics.
\newblock \emph{Phys. Rev.}, 150:\penalty0 1079--1085, Oct 1966.
\newblock \doi{10.1103/PhysRev.150.1079}.
\newblock URL \url{https://link.aps.org/doi/10.1103/PhysRev.150.1079}.

\bibitem[Anderson(1982)]{Anderson82}
Brian~D.O. Anderson.
\newblock Reverse-time diffusion equation models.
\newblock \emph{Stochastic Processes and their Applications}, 12\penalty0 (3):\penalty0 313--326, 1982.
\newblock ISSN 0304-4149.
\newblock \doi{https://doi.org/10.1016/0304-4149(82)90051-5}.
\newblock URL \url{https://www.sciencedirect.com/science/article/pii/0304414982900515}.

\bibitem[Haussmann and Pardoux(1986)]{Haussmann86}
U.~G. Haussmann and E.~Pardoux.
\newblock {Time Reversal of Diffusions}.
\newblock \emph{The Annals of Probability}, 14\penalty0 (4):\penalty0 1188 -- 1205, 1986.
\newblock \doi{10.1214/aop/1176992362}.
\newblock URL \url{https://doi.org/10.1214/aop/1176992362}.

\bibitem[F{\"o}llmer(1988)]{Follmer88}
Hans F{\"o}llmer.
\newblock Random fields and diffusion processes.
\newblock In Paul-Louis Hennequin, editor, \emph{{\'E}cole d'{\'E}t{\'e} de Probabilit{\'e}s de Saint-Flour XV--XVII, 1985--87}, pages 101--203, Berlin, Heidelberg, 1988. Springer Berlin Heidelberg.
\newblock ISBN 978-3-540-46042-8.

\bibitem[Premkumar(2024)]{Premkumar24a}
Akhil Premkumar.
\newblock Neural entropy, 2024.
\newblock URL \url{https://arxiv.org/abs/2409.03817}.

\bibitem[Sohl-Dickstein et~al.(2015)Sohl-Dickstein, Weiss, Maheswaranathan, and Ganguli]{Sohl-Dickstein15}
Jascha Sohl-Dickstein, Eric Weiss, Niru Maheswaranathan, and Surya Ganguli.
\newblock Deep unsupervised learning using nonequilibrium thermodynamics.
\newblock In Francis Bach and David Blei, editors, \emph{Proceedings of the 32nd International Conference on Machine Learning}, volume~37 of \emph{Proceedings of Machine Learning Research}, pages 2256--2265, Lille, France, 07--09 Jul 2015. PMLR.
\newblock URL \url{https://proceedings.mlr.press/v37/sohl-dickstein15.html}.

\bibitem[Maoutsa et~al.(2020)Maoutsa, Reich, and Opper]{Maoutsa20}
Dimitra Maoutsa, Sebastian Reich, and Manfred Opper.
\newblock Interacting particle solutions of fokker–planck equations through gradient–log–density estimation.
\newblock \emph{Entropy}, 22\penalty0 (8), 2020.
\newblock ISSN 1099-4300.
\newblock \doi{10.3390/e22080802}.
\newblock URL \url{https://www.mdpi.com/1099-4300/22/8/802}.

\bibitem[Yasue(1978)]{Yasue78}
Kunio Yasue.
\newblock {A simple derivation of the Onsager–Machlup formula for one‐dimensional nonlinear diffusion process}.
\newblock \emph{Journal of Mathematical Physics}, 19\penalty0 (8):\penalty0 1671--1673, 08 1978.
\newblock ISSN 0022-2488.
\newblock \doi{10.1063/1.523888}.
\newblock URL \url{https://doi.org/10.1063/1.523888}.

\bibitem[Seifert(2012)]{Seifert12}
Udo Seifert.
\newblock Stochastic thermodynamics, fluctuation theorems and molecular machines.
\newblock \emph{Reports on Progress in Physics}, 75\penalty0 (12):\penalty0 126001, nov 2012.
\newblock \doi{10.1088/0034-4885/75/12/126001}.
\newblock URL \url{https://dx.doi.org/10.1088/0034-4885/75/12/126001}.

\bibitem[Pavon(1989)]{Pavon89}
Michele Pavon.
\newblock Stochastic control and nonequilibrium thermodynamical systems.
\newblock \emph{Applied Mathematics and Optimization}, 19\penalty0 (1):\penalty0 187--202, 1989.
\newblock \doi{10.1007/BF01448198}.
\newblock URL \url{https://doi.org/10.1007/BF01448198}.

\bibitem[Huang et~al.(2021)Huang, Lim, and Courville]{Huang21}
Chin-Wei Huang, Jae~Hyun Lim, and Aaron~C Courville.
\newblock A variational perspective on diffusion-based generative models and score matching.
\newblock In M.~Ranzato, A.~Beygelzimer, Y.~Dauphin, P.S. Liang, and J.~Wortman Vaughan, editors, \emph{Advances in Neural Information Processing Systems}, volume~34, pages 22863--22876. Curran Associates, Inc., 2021.
\newblock URL \url{https://proceedings.neurips.cc/paper_files/paper/2021/file/c11abfd29e4d9b4d4b566b01114d8486-Paper.pdf}.

\bibitem[Ho et~al.(2020)Ho, Jain, and Abbeel]{Ho20}
Jonathan Ho, Ajay Jain, and Pieter Abbeel.
\newblock Denoising diffusion probabilistic models.
\newblock In Hugo Larochelle, Marc'Aurelio Ranzato, Raia Hadsell, Maria{-}Florina Balcan, and Hsuan{-}Tien Lin, editors, \emph{Advances in Neural Information Processing Systems 33: Annual Conference on Neural Information Processing Systems 2020, NeurIPS 2020, December 6-12, 2020, virtual}, 2020.
\newblock URL \url{https://proceedings.neurips.cc/paper/2020/hash/4c5bcfec8584af0d967f1ab10179ca4b-Abstract.html}.

\bibitem[Premkumar(2023)]{Premkumar23}
Akhil Premkumar.
\newblock Generative diffusion from an action principle, 2023.
\newblock URL \url{https://arxiv.org/abs/2310.04490}.

\bibitem[Durkan et~al.(2020)Durkan, Bekasov, Murray, and Papamakarios]{nflows}
Conor Durkan, Artur Bekasov, Iain Murray, and George Papamakarios.
\newblock {nflows}: normalizing flows in {PyTorch}, November 2020.
\newblock URL \url{https://doi.org/10.5281/zenodo.4296287}.

\bibitem[Fan et~al.(2013)Fan, Nott, and Sisson]{Fan12}
Yanan Fan, David~J. Nott, and Scott~A. Sisson.
\newblock Approximate bayesian computation via regression density estimation.
\newblock \emph{Stat}, 2\penalty0 (1):\penalty0 34--48, 2013.
\newblock \doi{https://doi.org/10.1002/sta4.15}.
\newblock URL \url{https://onlinelibrary.wiley.com/doi/abs/10.1002/sta4.15}.

\bibitem[Li et~al.(2023)Li, Prabhudesai, Duggal, Brown, and Pathak]{Li23}
Alexander~C. Li, Mihir Prabhudesai, Shivam Duggal, Ellis Brown, and Deepak Pathak.
\newblock Your diffusion model is secretly a zero-shot classifier.
\newblock In \emph{{IEEE/CVF} International Conference on Computer Vision, {ICCV} 2023, Paris, France, October 1-6, 2023}, pages 2206--2217. {IEEE}, 2023.
\newblock \doi{10.1109/ICCV51070.2023.00210}.
\newblock URL \url{https://doi.org/10.1109/ICCV51070.2023.00210}.

\bibitem[Vincent(2011)]{Vincent11}
Pascal Vincent.
\newblock A connection between score matching and denoising autoencoders.
\newblock \emph{Neural Computation}, 23:\penalty0 1661--1674, 2011.
\newblock URL \url{https://api.semanticscholar.org/CorpusID:5560643}.

\bibitem[Hutchinson(1990)]{Hutchinson90}
M.~F. Hutchinson.
\newblock A stochastic estimator of the trace of the influence matrix for laplacian smoothing splines.
\newblock \emph{Communications in Statistics - Simulation and Computation}, 19\penalty0 (2):\penalty0 433--450, 1990.
\newblock \doi{10.1080/03610919008812866}.
\newblock URL \url{https://doi.org/10.1080/03610919008812866}.

\bibitem[Adams et~al.(2018)Adams, Pennington, Johnson, Smith, Ovadia, Patton, and Saunderson]{Adams18}
Ryan~P. Adams, Jeffrey Pennington, Matthew~J. Johnson, Jamie Smith, Yaniv Ovadia, Brian Patton, and James Saunderson.
\newblock Estimating the spectral density of large implicit matrices, 2018.
\newblock URL \url{https://arxiv.org/abs/1802.03451}.

\end{thebibliography}

\end{document}